\documentclass{article} 
\usepackage{iclr2025_conference,times}

\usepackage{amsmath,amsfonts,bm}

\def\eqref#1{equation~\ref{#1}}

\def\1{\bm{1}}

\DeclareMathAlphabet{\mathsfit}{\encodingdefault}{\sfdefault}{m}{sl}
\SetMathAlphabet{\mathsfit}{bold}{\encodingdefault}{\sfdefault}{bx}{n}

\usepackage{hyperref}
\usepackage{amsfonts}
\usepackage{varioref}
\usepackage{cleveref}
\usepackage{url}
\usepackage[T1]{fontenc}
\usepackage{microtype}

\usepackage{inconsolata}

\usepackage{graphicx}

\usepackage[utf8]{inputenc} 
\usepackage[T1]{fontenc}    
\usepackage{url}            
\usepackage{booktabs}       
\usepackage{amsfonts}       
\usepackage{nicefrac}       
\usepackage{microtype}      
\usepackage{xcolor}         
\usepackage{amsmath}
\usepackage{xspace}
\usepackage{listings}
\usepackage{cleveref}
\usepackage{multirow}
\usepackage[colorinlistoftodos,prependcaption,textsize=tiny]{todonotes}
\usepackage{lipsum} 
\usepackage{xargs} 
\usepackage{enumerate}
\usepackage{tcolorbox}
\usepackage{times}
\usepackage{adjustbox}

\definecolor{leanblue}{RGB}{0,102,204}
\definecolor{leangreen}{RGB}{0,153,0}
\definecolor{leanpurple}{RGB}{102,0,204}
\definecolor{leandarkgray}{RGB}{64,64,64}
\definecolor{leanlightgray}{RGB}{192,192,192}
\definecolor{leanred}{RGB}{170, 0, 0}
\lstdefinelanguage{lean}{
    keywords=[1]{import,namespace,section,end,variable,variables,parameter,parameters,def,theorem,axiom,example,inductive,structure,class,universe,alias,notation,infixl,infixr,infix,prefix,postfix,reserve,theorem,axiom,lemma,corollary,variable,variables,parameter,parameters,definition,meta,check,theorem,example,proof,by,fun},
    keywordstyle=[1]\bfseries\color{leanblue},
    keywords=[2]{Prop,Type,Set,Type*,nat,int,bool,let,have,assume,show,from},
    keywordstyle=[2]\color{leangreen},
    keywords=[3]{forall,exists,push_cast,pop_cast,ring_exp,linarith,norm_num,by_contradiction,rewrite,apply,exact,refl,congr,symmetry,contradiction,induction,split,case,unfold,refine,odd_iff},
    keywordstyle=[3]\color{leanpurple},
    keywords=[4]{λ,∀,∃,¬,∧,∨,→,↔,≠,≤,≥,∈,∉,⊆,∅,∩,∪,∘},
    keywordstyle=[4]\color{leandarkgray},
    comment=[l]{--},
    commentstyle=\color{leanlightgray},
    string=[b]{"},
    stringstyle=\color{red},
    sensitive=true,
    breaklines=true,
    breakatwhitespace=true,
    tabsize=4,
    numberstyle=\tiny\color{gray},
    basicstyle=\small\ttfamily,
    showstringspaces=false,
    literate=
        {←}{{$\leftarrow$}}1
        {→}{{$\rightarrow$}}1
        {↔}{{$\leftrightarrow$}}1
        {∀}{{$\forall$}}1
        {∃}{{$\exists$}}1
        {∅}{{$\emptyset$}}1
        {λ}{{$\lambda$}}1
        {¬}{{$\neg$}}1
        {∧}{{$\wedge$}}1
        {∨}{{$\vee$}}1
        {≠}{{$\neq$}}1
        {≤}{{$\leq$}}1
        {≥}{{$\geq$}}1
        {⊆}{{$\subseteq$}}1
        {∈}{{$\in$}}1
        {∉}{{$\notin$}}1
        {∘}{{$\circ$}}1
}
\lstdefinestyle{lean}{
    language=lean,
    columns=fullflexible,
    numberstyle=\tiny\color{gray},
    basicstyle=\footnotesize\ttfamily,
    keywordstyle={[1]\bfseries\color{leanred}},
    commentstyle=\color{leanlightgray},
    stringstyle=\color{red},
    breaklines=true,
    breakatwhitespace=true,
    tabsize=4,
    showspaces=false,
    showstringspaces=false,
    literate=
        {←}{{$\leftarrow$}}1
        {→}{{$\rightarrow$}}1
        {↔}{{$\leftarrow\hspace{-1.0em}\rightarrow\hspace{0.2em}$}}1
        {∀}{{$\forall$}}1
        {∃}{{$\exists$}}1
        {∅}{{$\emptyset$}}1
        {λ}{{$\lambda$}}1
        {¬}{{$\neg$}}1
        {∧}{{$\wedge$}}1
        {∨}{{$\vee$}}1
        {≠}{{$\neq$}}1
        {≤}{{$\leq$}}1
        {≥}{{$\geq$}}1
        {⊆}{{$\subseteq$}}1
        {∈}{{$\in$}}1
        {∉}{{$\notin$}}1
        {∘}{{$\circ$}}1
        {≡}{{$\equiv$}}1
        {×}{{$\times$}}1
        {∞}{{$\infty$}}1
        {∑}{{$\sum$}}1
        {∏}{{$\prod$}}1
        {∖}{{$\setminus$}}1
        {⟨}{{$\langle$}}1
        {⟩}{{$\rangle$}}1
        {α}{{$\alpha$}}1
        {β}{{$\beta$}}1
        {γ}{{$\gamma$}}1
        {δ}{{$\delta$}}1
        {ε}{{$\epsilon$}}1
        {ζ}{{$\zeta$}}1
        {η}{{$\eta$}}1
        {θ}{{$\theta$}}1
        {ι}{{$\iota$}}1
        {κ}{{$\kappa$}}1
        {λ}{{$\lambda$}}1
        {μ}{{$\mu$}}1
        {ν}{{$\nu$}}1
        {ξ}{{$\xi$}}1
        {ο}{{$o$}}1
        {π}{{$\pi$}}1
        {ρ}{{$\rho$}}1
        {σ}{{$\sigma$}}1
        {τ}{{$\tau$}}1
        {υ}{{$\upsilon$}}1
        {φ}{{$\phi$}}1
        {χ}{{$\chi$}}1
        {ψ}{{$\psi$}}1
        {ω}{{$\omega$}}1
        {ℤ}{{$\mathbb{Z}$}}1
        {ℚ}{{$\mathbb{Q}$}}1
        {ℝ}{{$\mathbb{R}$}}1
        {₀}{{$_0$}}1
        {₁}{{$_1$}}1
        {₂}{{$_2$}}1
        {₃}{{$_3$}}1
        {₄}{{$_4$}}1
        {₅}{{$_5$}}1
        {₆}{{$_6$}}1
        {₇}{{$_7$}}1
        {₈}{{$_8$}}1
        {₉}{{$_9$}}1
}
\lstdefinestyle{json}{
  basicstyle=\ttfamily,
  language=Java,
  numbers=none,
  numberstyle=\tiny,
  numbersep=5pt,
  frae=single,
  framerule=0.5pt,
  rulecolor=\color{black!20},
  backgroundcolor=\color{gray!10},
  breaklines=true,
  breakatwhitespace=false,
  postbreak=\raisebox{0ex}[0ex][0ex]{\ensuremath{\color{red}\hookrightarrow\space}},
  showtabs=false,
  showspaces=false,
  showstringspaces=false,
  keywordstyle=\color{blue},
  stringstyle=\color{green!40!black},
  commentstyle=\color{red!50!green},
  morestring=[b]",
  morestring=[d]',
  morecomment=[s]{/*}{*/},
  morecomment=[l]//,
  morekeywords={true,false,null},
  sensitive=false,
  emph={string,number,array,object,true,false,null},
  emphstyle=\color{black}\bfseries
}

\lstdefinestyle{informal}{
    language=,
    basicstyle=\small\ttfamily,
    keywordstyle=\bfseries,
    commentstyle=\itshape,
    stringstyle=\color{red},
    breaklines=true,
    breakatwhitespace=true,
    columns=fullflexible,
    numbers=none,
    captionpos=b,
    frame=none,
    escapeinside={(*}{*)},
    literate=
        {`}{{\textquotesingle}}1
        {→}{{$\rightarrow$}}1
        {←}{{$\leftarrow$}}1
        {↔}{{$\leftrightarrow$}}1
        {≠}{{$\neq$}}1
        {≤}{{$\leq$}}1
        {≥}{{$\geq$}}1
        {⊆}{{$\subseteq$}}1
        {∈}{{$\in$}}1
        {∉}{{$\notin$}}1
        {∘}{{$\circ$}}1
        {↑}{{$\uparrow$}}1
        {π}{{$\pi$}}1
        {⊢}{{$\vdash $}}1
}

\definecolor{mybrown}{RGB}{128,64,0}
\gdef\Sepline{%
  \par\noindent\makebox[\linewidth][l]{%
  \hspace*{-\mdflength{innerleftmargin}}%
   \tikz\draw[thick,dashed,gray!60] (0,0) --%
        (\textwidth+\the\mdflength{innerleftmargin}+\the\mdflength{innerrightmargin},0);
  }\par\nobreak}

\definecolor{isarblue}{HTML}{006699}
\definecolor{isarfaintblue}{rgb}{0.0, 0.75, 1.0}
\definecolor{isargreen}{HTML}{009966}
\definecolor{red}{HTML}{990000}
\definecolor{patriarch}{rgb}{0.5, 0.0, 0.5}
\lstdefinelanguage{isabelle}{%
    keywords=[1]{type_synonym,datatype,fun,abbreviation,definition,proof,lemma,theorem,qed,corollary,have,hence,also,finally,ultimately,moreover,using,\{},
    keywordstyle=[1]\bfseries\color{isarblue},
    keywords=[2]{where,assumes,shows,fixes,and},
    keywordstyle=[2]\bfseries\color{isargreen},
    keywords=[3]{if,then,else,case,SOME,let,in,O},
    keywordstyle=[3]\color{isarblue},
    keywords=[4]{ATP},
    keywordstyle=[4]\it\color{patriarch},
    keywords=[5]{show,assume,obtain},
    keywordstyle=[5]\bfseries\color{isarfaintblue},
}

\lstdefinestyle{isabelle}{%
  language=isabelle,
  escapeinside={\&}{&},
  columns=fixed,
  extendedchars,
  basewidth={0.5em,0.45em},
  basicstyle=\singlespacing\ttfamily\small,
  mathescape,
  morecomment=[s][\bfseries\color{red}]{(*}{*)},
  morecomment=[l][\bfseries]{####},
}

\lstdefinelanguage{json}{
  basicstyle=\normalfont\ttfamily,
  numbers=left,
  numberstyle=\scriptsize,
  stepnumber=1,
  numbersep=8pt,
  showstringspaces=false,
  breaklines=true,
  frame=lines,
  backgroundcolor=\color{background},
  literate=
  *{0}{{{\color{numb}0}}}{1}
   {1}{{{\color{numb}1}}}{1}
   {2}{{{\color{numb}2}}}{1}
   {3}{{{\color{numb}3}}}{1}
   {4}{{{\color{numb}4}}}{1}
   {5}{{{\color{numb}5}}}{1}
   {6}{{{\color{numb}6}}}{1}
   {7}{{{\color{numb}7}}}{1}
   {8}{{{\color{numb}8}}}{1}
   {9}{{{\color{numb}9}}}{1}
   {:}{{{\color{punct}{:}}}}{1}
   {,}{{{\color{punct}{,}}}}{1}
   {\{}{{{\color{delim}{\{}}}}{1}
   {\}}{{{\color{delim}{\}}}}}{1}
   {[}{{{\color{delim}{[}}}}{1}
   {]}{{{\color{delim}{]}}}}{1},
   {↑}{{$\uparrow$}}1
   {π}{{$\pi$}}1
   {⊢}{{$\Longrightarrow $}}1
}

\newcommand{\yinya}[1]{\textcolor{black}{#1}}

\newcommand{\method}{\textsc{FormalAlign}\xspace}
\newcommand{\model}{\method model\xspace}
\newcommand{\forml}{FormL4\xspace}

\definecolor{lightred}{RGB}{255,204,204}

\newcommandx{\info}[2][1=]{\todo[lineco lor=red,backgroundcolor=red!25,bordercolor=red,#1]{#2}}

\title{\method: Automated Alignment \\ Evaluation for Autoformalization}

\iclrfinalcopy

\author{\textbf{Jianqiao Lu}$^1$\thanks{Leading co-authors with equal contribution.}\hspace{4px}, \textbf{Yingjia Wan}$^{2*}$\textbf{,} 
\textbf{Yinya Huang}$^{4}$\textbf{,} 
\textbf{Jing Xiong}$^{1}$\textbf{,}
 \textbf{Zhengying Liu}$^{3}$\textbf{,}   
\textbf{Zhijiang Guo}$^3$\thanks{Corresponding Author.}
\\
$^1$The University of Hong Kong \ \ \ $^2 $University of Cambridge    \\ 
$^3$Huawei Noah’s Ark Lab \ \ \ $^4 $City University of Hong Kong \\ 
\texttt{jqlu@cs.hku.hk, \{yingjiawan.alisa, cartusguo\}@gmail.com}\\
}

\begin{document}

\maketitle

\begin{abstract}


Autoformalization aims to convert informal mathematical proofs into machine-verifiable formats, bridging the gap between natural and formal languages. However, ensuring semantic alignment between the informal and formalized statements remains challenging. Existing approaches heavily rely on manual verification, hindering scalability. 
To address this, we introduce \method, the first automated framework designed for evaluating the alignment between natural and formal languages in autoformalization. \method trains on both the autoformalization sequence generation task and the representational alignment between input and output, employing a dual loss that combines a pair of mutually enhancing autoformalization and alignment tasks. Evaluated across four benchmarks augmented by our proposed misalignment strategies, \method demonstrates superior performance. In our experiments, \method outperforms GPT-4, achieving an Alignment-Selection Score 11.58\% higher on \forml-Basic (99.21\% vs. 88.91\%) and 3.19\% higher on MiniF2F-Valid (66.39\% vs. 64.34\%). 
This effective alignment evaluation significantly reduces the need for manual verification. 
Both the dataset and code can be accessed via~\url{https://github.com/rookie-joe/FormalAlign}.

\end{abstract}

\section{Introduction}
\label{sec:intro}
Autoformalization is the task of automatically converting informal theorems and proofs into machine-verifiable formats~\citep{WangKU18,Szegedy20, Yuhuai2022nips, JiangWZL0LJLW23}. It bridges the gap between natural and formal languages, leveraging the strengths of both: natural language carries extensive logical reasoning and human knowledge. while formal language enables rigorous verification and proof~\citep{Kaliszy2014Corr}.  While promising, autoformalization faces challenges in ensuring semantic alignment between these languages. 
The availability of fully formalized and computer-checked content is limited~\citep{KaliszykUV17}. This lack of alignment information hinders the development of robust autoformalization models~\citep{Bansal2020AITP}.

\begin{figure}[ht]
    \centering
    \vspace{-2em}
    \includegraphics[width=\columnwidth]{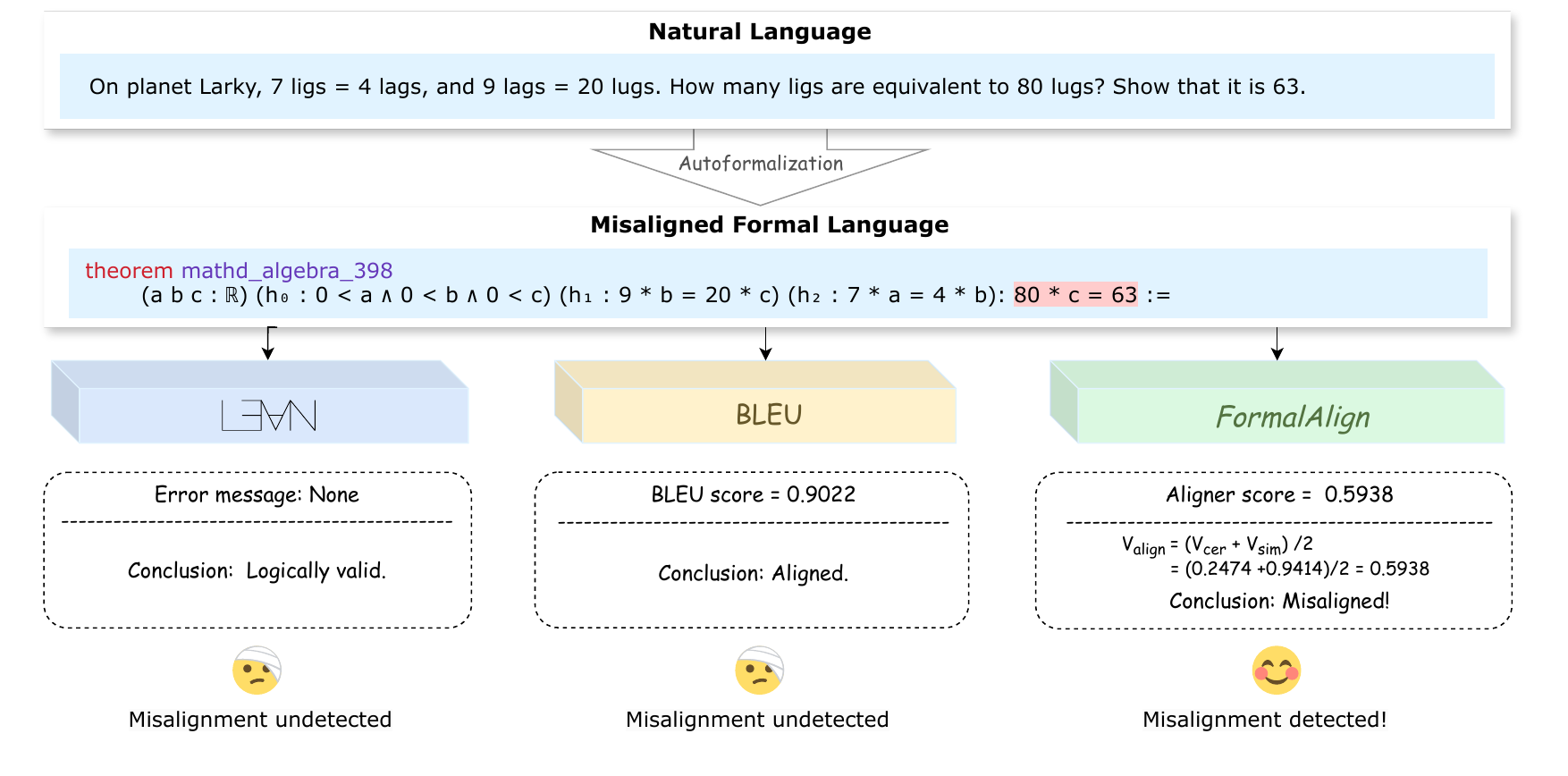}
    \vspace{-2em}
    \caption{
    A comparison of current methods and \method in evaluating autoformalization. The formal statement is misaligned with the natural language statement: it incorrectly ends with
    \colorbox{lightred}{\(80 * c = 63\)}, when the aligned equation should be \(63 * a = 80 * c\). Current methods can only verify the surface-form integrity of the autoformalized sequence via BLEU or by passing it to a formal language compiler, while our \method successfully detects the semantic misalignment of the autoformalized statement with the informal sequence.}
    \label{fig:model-example}
\end{figure}

Current evaluation for autoformalization \citep{JiangWZL0LJLW23,huang2024mustard} focus solely on logical validity, which can be easily verified by formal language compilers  (e.g., the Lean 4 compiler\footnote{Details of the compiler are provided in Appendix~\ref{app:compiler}.}). 
Another direct but suboptimal evaluation resort to surface form matching via BLEU~\citep{DBLP:conf/acl/PapineniRWZ02}, which is widely used by recent works~\citep{Yuhuai2022nips, JiangWZL0LJLW23, AzerbayevProofNet2023}, but struggles with semantic alignment or logical equivalence~\citep{Li2024Corr}.

Take the case in Figure~\ref{fig:model-example} as an example,
to correctly translate the natural language proof target into a Lean 4 statement, 
first, the variables for the objects ``ligs'', ``lags'', and ``lugs'' should be included and real numbers greater than zero.
Then, the two equations should be translated into two corresponding hypotheses $h_1$ and $h_2$.
Finally, the proof target ``How many ligs are equivalent to 80 lugs? Show that 63'' needs to be formalized into ``$63 * a = 80 * c$'', which is failed in this case by omitting the pronounced ``ligs''.
However, because the incorrectly translated ``$80 * c = 63$'' is logically valid in Lean 4 and similar to the ground truth in surface form, it is flawless to a theorem compiler
or the BLEU score. The semantic misalignment of lacking a ``lig'' in the equation is undetected. 
Moreover, due to its elusive nature, this misalignment is often challenging to detect, even with methods like BERTscore \citep{zhang2020bertscore}, which are designed to assess semantic similarity, 
Therefore, a robust and effective approach to Automated Alignment Evaluation (AAE) is urgently needed.

\begin{figure*}[!t]
    \centering
    \includegraphics[trim=0 0 75 0,clip,width=\textwidth]{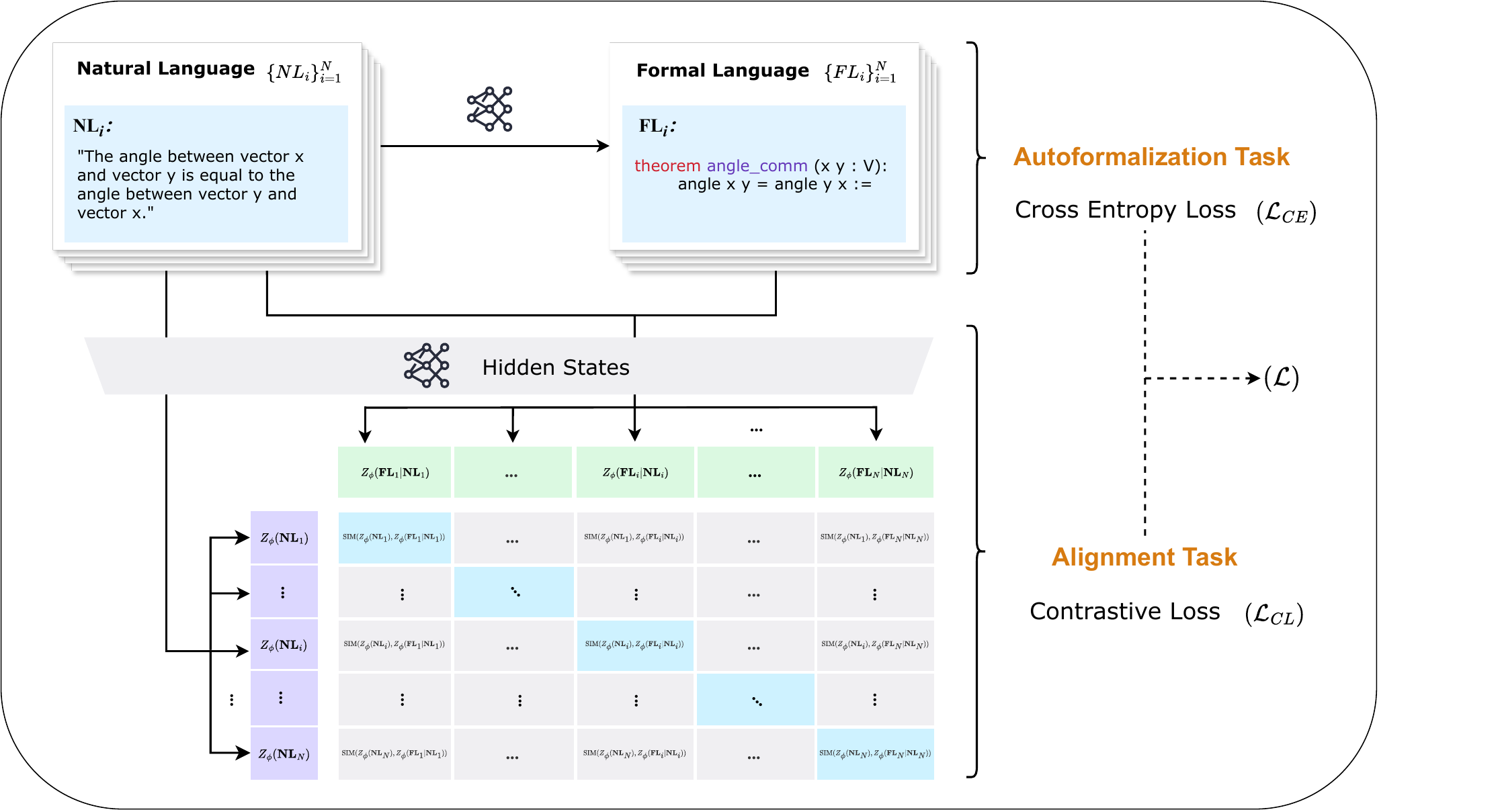}
    \caption{An overview of \method, which combines the cross-entropy loss in sequence autoformalization and the contrastive loss in hidden states to enhance the informal-formal alignment.}
    \label{fig:method-overview}
\end{figure*}

To bridge this gap, we introduce the \textbf{\method} framework, which 
assesses the alignment between informal and formal languages during autoformalization. As demonstrated in Figure~\ref{fig:method-overview},
\method learns both the sequence generation task of autoformalization \yinya{(top half in Figure~\ref{fig:method-overview})} and the representational alignment \yinya{(bottom half in Figure~\ref{fig:method-overview})} between input and output. 
\method jointly trains the pair of mutually enhancing tasks.
This encourages the model to generate similar embeddings for corresponding pairs and distinct embeddings for non-corresponding pairs, enhancing its ability to differentiate between aligned and misaligned sequences,
\yinya{as the case in Figure \ref{fig:model-example}.}

We evaluate \method on four benchmarks sourced from MiniF2F~\citep{ZhengHP22} and \forml~\citep{Lu2024PDA}.
Compared with GPT-4, \method achieves a substantially higher precision score across these datasets, e.g., in the \forml-Basic dataset (93.65\% vs. 26.33\%). 
It also outperforms GPT-4 in alignment-selection score across multiple datasets, including a remarkable 99.21\% vs. 88.91\% in \forml-Basic and 66.39\% vs. 64.34\% in MiniF2F-Valid.
Extensive experimental results demonstrate the effectiveness of \method, significantly reducing the reliance on manual verification. 
Our contributions are summarized as follows:

\begin{itemize}
\item To the best of our knowledge, we design the \textbf{first} method for automatically evaluating alignment in autoformalization, reducing the reliance for manual verification.
\item We develop a combined loss framework that simultaneously enhances a model for both autoformalization and semantic alignment.
\item Extensive experiments on established autoformalization benchmarks demonstrate the effectiveness and robustness of \method. 
\end{itemize}

\section{Related Work}

\paragraph{Autoformalization with LLMs}
Early efforts~\citep{WangKU18,Bansal2020AITP} employed encoder-decoder neural networks to translate informal statements into formal languages like Mizar~\citep{Rudnicki1992AnOO}, HOL Light~\citep{Harrison1996Hollight}, and Coq~\citep{Barras1997Coq}.
The advent of LLMs~\citep{Chen2021Corr,Chowdhery2022PaLM,Lewkowycz2022Minerva,GPT4} enhances the capabilities of autoformalization~\citep{Yuhuai2022nips,JiangMMA2023}. 
Some approaches  directly prompt LLMs~\citep{Yuhuai2022nips,JiangWZL0LJLW23,Zhao2023Corr,JiangMMA2023} to translate mathematical problems into formal languages like Isabelle~\citep{MakariusIsabelle2008} and Lean~\citep{Leonardo2015Lean3}. 
On the other hand, training or fine-tuning LLMs with paired formal-informal data~\citep{AzerbayevLLemma2023,Huaiyuan2024InternLM-Math,Zhihong2024Deepseek-math,Lu2024PDA} garner increasing attention for its effectiveness in enhancing LLMs' performance in autoformalization.
The evaluation of autoformalization primarily depends on manually verifying the alignment between informal and formalized statements~\citep{Li2024Corr}. 
There is a pressing need for an efficient and less labor-intensive method for automated autoformalization alignment.

\paragraph{Evaluation for Generation}
The challenges of automated evaluating natural language generation tasks, grow as the difficulty of tasks increases. 
N-gram-based metrics \citep{DBLP:conf/acl/PapineniRWZ02,lin-2004-rouge} resort to surface-form matching, which has been beneficial for evaluation tasks with static references such as image-captioning \citep{DBLP:journals/tacl/YoungLHH14,DBLP:journals/corr/ChenFLVGDZ15} and text summarization \citep{SeeLM17,NarayanCL18}. 
Semantics are rarely considered until embedding-based metrics emerge, especially metrics leveraging the evolving pre-trained language models \citep{DBLP:conf/iclr/ZhangKWWA20,BARTScore2021nips,T5Score2023EMNLP} 
and LLMs \citep{Xu2023EMNLP, Gpteval2023arxiv, Liu2024Align}.
The growth of LLMs continues empowering parameter-based metrics for advanced evaluation such as multi-agent \citep{ChatEval2023Corr}, multi-aspect \citep{X-Eval2023Corr} and multi-proxies~\citep{Tan2024Proxy}
%
%
The other line of work fine-tunes language models for scoring \citep{Ke2023Corr,Wang2023Corr,Yue2023EMNLP}, labeling \citep{Gekhman2023ACL}, text probability calculation \citep{X-Eval2023Corr}, or comparison \citep{Wang2023Corr,Zheng2023nips} to enhance and adjust for evaluation targets. In this paper, we propose an automated evaluator for the challenging yet under-explored autoformalization evaluation that requires both rigorous logical validity and aligned semantics between the natural-formal pair. 
To this end, we fine-tune LLMs via joint autoformalization generation and representational alignment tasks and obtain a logically and semantically empowered aligner.




\section{Method: \method}
\label{sec:method}

The \textbf{\method} framework trains an LLM that can evaluate the alignment between natural (informal) and formal languages during autoformalization. As illustrated in Figure \ref{fig:method-overview}, \method combines two types of loss in the training process: one for the sequence generation task of autoformalization and another for the representational alignment between input and output. This dual loss framework mutually enhances autoformalization and alignment.

\subsection{Notations}
We first define the notations as follows:

\noindent $\mathbf{NL}_i$ : The $i^{th}$  informal input sequence in a batch, $\mathbf{NL}_i = (\mathbf{NL}_{i,1}, \mathbf{NL}_{i,2}, \ldots, \mathbf{NL}_{i,m})$ , where $m$ is the sequence length of $\mathbf{NL}_i$ .

\noindent $\mathbf{FL}_i$ : The $i^{th}$ ground-truth formal output sequence in a batch, $\mathbf{FL}_i = (\mathbf{FL}_{i,1}, \mathbf{FL}_{i,2}, \ldots, \mathbf{FL}_{i,n})$ , where $n$ is the sequence length of $\mathbf{FL}_i$ .

\noindent $P_{\phi}(\mathbf{FL}_{i,j} | \mathbf{FL}_{i,<j}, \mathbf{NL}_i)$ : The probability of predicting the $j^{th}$ token in the formal sequence $\mathbf{FL}_i$ by the auto-regressive language model with parameters $\phi$ , given the previous tokens $\mathbf{FL}_{i,<j}$ in the formal sequence and the informal input $\mathbf{NL}_i$ .

\noindent $Z_{\phi}(\mathbf{NL}_i)$ : The hidden state from the auto-regressive language model with parameters $\phi$ for the final position in the $i^{th}$ informal input $\mathbf{NL}_i$ , i.e., $\mathbf{NL}_{i,m}$ .

\noindent $Z_{\phi}(\mathbf{FL}_i | \mathbf{NL}_i)$ : The hidden state from the auto-regressive language model with parameters $\phi$ for the final position in the $i^{th}$ ground-truth formal output $\mathbf{FL}_i$ , i.e., $\mathbf{FL}_{i,n}$ , conditioned on the paired $i^{th}$  informal input $\mathbf{NL}_i$ .

\noindent  $Z_{\phi}(\mathbf{FL}_{i'} | \mathbf{NL}_i)$ : The hidden state from the auto-regressive language model with parameters $\phi$ for the final position in the $(i')^{th}$ unpaired formal output $\mathbf{FL}_{i'}$ in a batch, conditioned on the $i^{th}$ informal input $\mathbf{NL}_i$ .

\noindent  $\text{cos}(\cdot, \cdot)$ : The cosine similarity between embeddings, defined as \(\text{cos}(x, y) = \frac{x \cdot y}{\|x\| \cdot \|y\|}\).

\noindent  $N$ : The batch size.

\subsection{Training}

\paragraph{Autoformalization Task}

For the autoformalization task of converting an informal input sequence \( \mathbf{NL}_i \) to a formal output sequence \( \mathbf{FL}_i \), we use the cross-entropy loss function. This function measures the error in predicting each word in the formal sequence given the previous words and the informal input. It is defined as:
\begin{align*}
\mathcal{L}_{CE} = - \sum_{j=1}^{n} \log P_{\phi}(\mathbf{FL}_{i,j} | \mathbf{FL}_{i,j'| j' < j} , \mathbf{NL}_i) 
\end{align*}

\paragraph{Alignment Task}
To ensure that the embeddings of the informal and formal sequences are well-aligned in the \model, we introduce a contrastive loss $\mathcal{L}_{\text{CL}}$ . Let $\mathbf{u}_i$ and $\mathbf{v}_i$ denote the hidden state representations of the $i$ -th informal input $\mathbf{NL}_i$ and its corresponding formal output $\mathbf{FL}_i$ , respectively \(\mathbf{u}_i = Z_{\phi}(\mathbf{NL}_i)\) and  \(\mathbf{v}_i = Z_{\phi}(\mathbf{FL}_i | \mathbf{NL}_i) \).

The contrastive loss encourages the cosine similarity $\text{cos}(\mathbf{u}_i, \mathbf{v}_i)$ between the representations of corresponding informal-formal pairs to be higher than the cosine similarity $\text{cos}(\mathbf{u}_i, \mathbf{v}_{i'})$ between non-corresponding pairs:
\begin{align}
\label{eq: contrastive_loss}
\mathcal{L}_{CL} = - \frac{1}{N} \sum_{i=1}^N \log \frac{\exp\left({\text{cos}(\mathbf{u}_i, \mathbf{v}_i)}/{\tau}\right)}{\sum_{j=1}^N \exp\left({\text{cos}(\mathbf{u}_i, \mathbf{v}_j)}/{\tau}\right)}
\end{align}
where $\tau$ is a temperature parameter that scales the cosine similarities. 
By minimizing this contrastive loss, the \model learns to align the embeddings of corresponding informal-formal sequences while ensuring that the embeddings of non-corresponding sequences are dissimilar.




\paragraph{\method Loss}

We jointly train an evaluator model with the autoformalization and alignment tasks, resulting in a \model. The combined training loss is:

\begin{align}
\label{eq:main_loss}
\begin{aligned}
 \mathcal{L} = \mathcal{L}_{CE} + \mathcal{L}_{CL} 
\end{aligned}
\end{align}
We train an alignment-aware \model by minimizing a combined loss, enabling it to benefit from both the sequence alignment inherent in the autoformalization and the representation alignment facilitated by the contrastive learning process.

\subsection{Inference}
During the inference phase, the \model generates an alignment evaluation score $\mathcal{V}_{\text{align}}$ for each pair of informal input $\mathbf{NL}_i$ and formal output $\mathbf{FL}_i$ . 
This score combines two metrics: the certainty score and the similarity score.

\paragraph{Certainty Score}
The certainty score $\mathcal{V}_{\text{cer}}$ measures the confidence of the fine-tuned \model in predicting the formal output based on the corresponding informal input. It is calculated by taking the exponential of the average log-probability assigned by the model to each token in the formal sequence:
\begin{align}
\label{eq: certainty_score}
\mathcal{V}_{\text{cer}} = \exp\left (\frac{1}{n} \sum_{j=1}^n \log P_{\phi}(\mathbf{FL}_{i, j} | \mathbf{FL}_{i,<j}, \mathbf{NL}_i)\right)
\end{align}
where $P_{\phi}$ represents the probability output of the model with parameters $\phi$ , $\mathbf{FL}_{i,<j}$ denotes the tokens in the formal sequence up to position $j-1$ , and $n$ is the length of the formal sequence.

\paragraph{Similarity Score}
The similarity score $\mathcal{V}_{\text{sim}}$ measures alignment between the embedding representations of the informal input and the formal output. It is computed using the cosine similarity between the hidden states of the informal input and the formal output conditioned on the informal input:
\begin{align}
\label{eq: similarity_score}
\mathcal{V}_{\text{sim}} = \text{cos}(Z_{\phi}(\mathbf{NL}_i), Z_{\phi}(\mathbf{FL}_i | \mathbf{NL}_i))
\end{align}
where $Z_{\phi}(\mathbf{NL}_i)$ represents the hidden state from the final position in the informal input, and $Z_{\phi}(\mathbf{FL}_i | \mathbf{NL}_i)$ represents the hidden state from the formal output conditioned on informal input.
 

\paragraph{Alignment Score}
The overall alignment evaluation score $\mathcal{V}_{\text{align}}$ is computed by taking the average of the certainty score and the similarity score:
\begin{align}
\label{eq: alignment_score}
\mathcal{V}_{\text{align}} = {\left(\mathcal{V}_{\text{cer}} + \mathcal{V}_{\text{sim}}\right)}/{2}
\end{align}
This combined score reflects both the accuracy of the translation from informal to formal expressions and the alignment of the internal representations of the sequences, providing a robust evaluation metric during the inference stage.


\section{Experiment}
\subsection{Datasets}
\label{sec:datasets}

In our experimental setup, we conduct fine-tuning on the FormL4~\citep{Lu2024PDA} and MMA~\citep{JiangMMA2023} training sets, both of which are derived from Mathlib, a library of fundamental mathematical statements.  This training data enables our model to align informal mathematical statements with their formal counterparts.

We employ a comprehensive set of test sets that covers both in-domain and out-of-domain data. Specifically, we use four distinct test sets: the basic and random test sets from \forml, and the valid and test sets from MiniF2F~\citep{Zheng2022Minif2f}.
\forml, designed to assess the autoformalization capabilities of LLMs in Lean 4~\citep{LeonardoLean42021} sourced from Mathlib, provides a comprehensive evaluation framework. 
The basic and random test sets from \forml allow us to gauge the model's performance in autoformalizing fundamental math statements that are similar to the training data. 
In contrast, the validation and test sets from MiniF2F serve as out-of-domain test data, providing a more challenging evaluation setting. MiniF2F is a benchmark containing 488 manually formalized mathematical competition statements sourced from various mathematical olympiads (AMC, AIME, IMO) and high-school and undergraduate math classes. . 

\begin{figure*}[t]
    \centering
    \includegraphics[width=\textwidth]{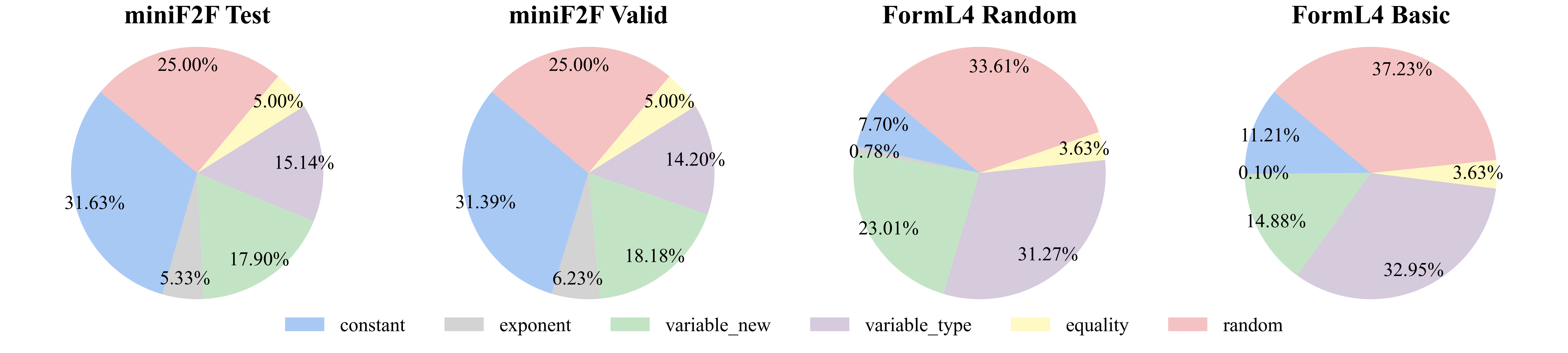}
     \caption{Distribution of misalignment types across datasets. This figure illustrates the variety and proportion of misalignment strategies applied to generate negative examples in the \forml-Basic, \forml-Random, MiniF2F-Valid, and MiniF2F-Test datasets.}
    \label{fig:misalignment_distribution}
\end{figure*}

These datasets primarily provide paired input-output instances, lacking the negative examples crucial for a more robust assessment of our model.  
Consider one aligned informal-formal pair shown in \Cref{tab:align_example} as an example.
We detail our approach to generating misaligned formal outputs with the natural (informal) input employing strategies outlined in \Cref{tab:misalignment_strategies}. The distribution of these misalignment types is visualized in~\Cref{fig:misalignment_distribution}. 
\begin{table}[h]
    \centering
    \caption{Natural Language Statement and its aligned Lean Formal Statement.}
    \vspace{3pt}
    \begin{tabular}{p{0.95\columnwidth}}
        \toprule
        \textbf{Natural Language Statement} \\
        \midrule
        The volume of a cone is given by the formula $V = \frac{1}{3}Bh$, where $B$ is the area of the base and $h$ is the height. The area of the base of a cone is 30 square units, and its height is 6.5 units. What is the number of cubic units in its volume? Show that it is 65. \\
        \midrule
        \textbf{Lean Formal Statement} \\
        \midrule
 \begin{lstlisting}[style=lean]
theorem mathd_algebra_478
  (b h v : ℝ)
  (h₀ : 0 < b ∧ 0 < h ∧ 0 < v)
  (h₁ : v = 1 / 3 * (b * h))
  (h₂ : b = 30)
  (h₃ : h = 13 / 2) :
  v = 65 :=
\end{lstlisting} \\
       \bottomrule
    \end{tabular}
    \vspace{-1.5em}
    \label{tab:align_example}
\end{table}

\begin{table}[h]
\centering
\caption{Misalignment strategies.}
\vspace{3pt}
\begin{tabular}{p{0.95\textwidth}}
\toprule
\begin{tabular}{p{0.45\textwidth}p{0.45\textwidth}}
\textbf{Constant Modification (constant)} & \textbf{Exponent Modification (exponent)} \\
This type of misalignment involves changing a constant value within the expression.
 \begin{lstlisting}[style=lean]
 theorem mathd_algebra_478
  (b h v : ℝ)
  (h_0 : 0 < b ∧ 0 < h ∧ 0 < v)
  (h_1 : v = 1 / 3 * (b * h))
  (h_2 : b = 31) -- changed constant
  (h_3 : h = 13 / 2) :
  v = 65 :=
 \end{lstlisting} &
This misalignment targets the exponents in the expression.
\begin{lstlisting}[style=lean]
theorem mathd_algebra_478
  (b h v : ℝ)
  (h_0 : 0 < b ∧ 0 < h ∧ 0 < v)
  (h_1 : v = 1 / 3 * (b^2 * h)) -- changed exponent
  (h_2 : b = 30)
  (h_3 : h = 13 / 2) :
  v = 65 :=
\end{lstlisting} \\
\end{tabular} \\

\begin{tabular}{p{0.45\textwidth}p{0.45\textwidth}}
\textbf{Introduction of a New Variable (variable\_new)} & \textbf{Change of Variable Type (variable\_type)} \\
This misalignment introduces a completely new variable into the expression. 
\begin{lstlisting}[style=lean]
theorem mathd_algebra_478
  (b h v x : ℝ) -- added a new variable x
  (h_0 : 0 < b ∧ 0 < h ∧ 0 < v)
  (h_1 : v = 1 / 3 * (b * h))
  (h_2 : b = 30)
  (h_3 : h = 13 / 2) :
  v = 65 :=
\end{lstlisting} &
In this case, the misalignment involves changing the type of a variable within the expression. The function identifies the type of a randomly selected variable and changes it to a different type from a predefined list of types.
\begin{lstlisting}[style=lean]
theorem mathd_algebra_478
  (b h v : ℚ) -- changed type to ℚ
  (h_0 : 0 < b ∧ 0 < h ∧ 0 < v)
  (h_1 : v = 1 / 3 * (b * h))
  (h_2 : b = 30)
  (h_3 : h = 13 / 2) :
  v = 65 :=
\end{lstlisting} \\
\end{tabular} \\

\begin{tabular}{p{0.45\textwidth}p{0.45\textwidth}}
\textbf{Modification of Equality (equality)} & \textbf{Random Pairing (random)} \\
This misalignment switches between equality \(=\) and inequality \(\neq\) symbols within the expression. 
\begin{lstlisting}[style=lean]
theorem mathd_algebra_478
  (b h v : ℝ)
  (h_0 : 0 < b ∧ 0 < h ∧ 0 < v)
  (h_1 : v ≠ 1 / 3 * (b * h)) -- swapped inequality
  (h_2 : b = 30)
  (h_3 : h = 13 / 2) :
  v = 65 :=
\end{lstlisting} &
This creates a mismatch between the informal input and its formal output. Instead of pairing the informal input with its correct formal output, this strategy randomly selects a formal output from other examples. \\
\end{tabular} \\
\bottomrule
\end{tabular}
\vspace{-1.5em}
\label{tab:misalignment_strategies}
\end{table}

\subsection{Metrics}
To assess the performance of models in evaluating the alignment of informal and formal language pairs, we introduce three automated metrics:

\textbf{Alignment Selection (AS):}
This metric quantifies how well a model selects the aligned formal output from multiple candidates when given an informal input. We calculate the alignment evaluation score $\mathcal{V}_{\text{align}}$ as described in~\Cref{sec:method} for each informal-formal pair. The pair with the highest score is selected as the aligned pair.

\textbf{Alignment Detection:}
We introduce a predefined threshold $\theta$ to detect the alignment for each informal-formal pair. If $\mathcal{V}_{\text{align}}$ exceeds $\theta$ , the model considers the pair to be aligned. We evaluate this detection method using two metrics: precision and recall. \textbf{Precision} measures the fraction of pairs identified as aligned by the model that is truly informal-formal pairs. It is calculated as \(\text{Precision} = \frac{TP}{TP + FP}\), where $TP$ represents the number of true positives (correctly identified aligned pairs) and $FP$ represents the number of false positives (incorrectly identified aligned pairs). \textbf{Recall} measures the fraction of true informal-formal pairs correctly identified by the model. It is calculated as: \(\text{Recall} = \frac{TP}{TP + FN}\), where $FN$ represents the number of false negatives (missed aligned pairs).







\begin{table*}[h]
\centering
\caption{Automated Alignment Evaluation (AAE) results across different benchmarks. The table compares the performance of our fine-tuned Mistral-7B model (\model) with GPT-4 and GPT-3.5 on four datasets: \forml-Basic, \forml-Random, MiniF2F-Valid, and MiniF2F-Test. Performance metrics include Alignment Score (AS), Precision (Prec.), and Recall (Rec.).}
\vspace{3pt}
\resizebox{\textwidth}{!}{
\begin{tabular}{lcccccccccccc}
\toprule
\multirow{2}{*}{\textbf{ Datasets}}     & \multicolumn{3}{c}{ \textbf{\forml-Basic} } & \multicolumn{3}{c}{\textbf{\forml-Random}} & \multicolumn{3}{c}{\textbf{MiniF2F-Valid}}   & \multicolumn{3}{c}{\textbf{MiniF2F-Test}}   \\ 
\cmidrule(lr){2-4} \cmidrule(lr){5-7}  \cmidrule(lr){8-10}  \cmidrule(lr){11-13} & \textbf{AS} & \textbf{Prec.} & \textbf{Rec.} & \textbf{AS} & \textbf{Prec.} & \textbf{Rec.} &  \textbf{AS} & \textbf{Prec.} & \textbf{Rec.} &  \textbf{AS} & \textbf{Prec.} & \textbf{Rec.}  \\
\midrule
GPT-4  & 88.91 & 26.33 & 88.69 & \textbf{90.52} & 28.56 & \textbf{90.02} & 64.34 & 44.58  & \textbf{90.98} & \textbf{68.31} & 51.11 & \textbf{94.65}  \\
GPT-3.5 & 50.23 &25.21 & \textbf{90.83} & 47.00 &  23.42  & 67.26 & 47.32 & 22.29 & 62.55 & 40.74 & 21.97 &  61.73   \\
\method & \textbf{99.21}& \textbf{93.65}& 86.43& 85.85& \textbf{86.90}& 89.20 &\textbf{66.39}& \textbf{68.58} & 60.66  &64.61 &\textbf{66.70 }& 63.37 \\
\bottomrule
\end{tabular}
}
\label{tab:main_table}
\vspace{-0.5em}
\end{table*}

\subsection{Main Results}

We fine-tune a Mistral-7B model~\citep{Albert2023Mistral} as the \model and evaluate its performance on various autoformalization benchmarks. The datasets used in this study include \forml-Basic, \forml-Random, MiniF2F-Valid, and MiniF2F-Test. Each data example consists of an aligned informal-formal pair, which is considered a positive example. To comprehensively assess the model's performance and robustness, we augment each positive example with 21 negative examples generated through carefully designed misalignment strategies outlined in~\Cref{tab:misalignment_strategies}. 

To balance precision and recall in the \model's alignment detection, we set \(\theta=0.7\). \Cref{tab:main_table} presents the detailed experimental results, including Alignment-Selection (AS), Precision (Prec.), and Recall (Rec.) metrics. The table compares the performance of our fine-tuned Mistral-7B model (\model) with GPT-4~\citep{GPT4} and GPT-3.5~\citep{GPT35turbo} across the different datasets. For more information on the query prompts used in the experiments, please refer to \Cref{app:prompt-details}.




\noindent \textbf{Effective and Robust Alignment Evaluation:} The experimental results demonstrate the effectiveness and robustness of our \method in evaluating the alignment between informal and formal languages. The model achieves impressive performance, with high alignment, precision, and recall scores across all datasets. Notably, on the \forml-Basic dataset, it attains an exceptional Alignment-Selection score of 99.21\% and a Precision of 93.65\%. These results highlight the model's ability to accurately identify aligned informal-formal pairs.

\noindent \textbf{Generalization Across Datasets:} 
The \model exhibits consistent performance across four diverse datasets, demonstrating its ability to generalize its autoformalization evaluation capabilities. Particularly noteworthy are the model's AS scores of 66.39\% and 64.61\% on the challenging MiniF2F-Valid and MiniF2F-Test datasets, respectively. These scores are comparable to those achieved by GPT-4, which obtained AS scores of 64.34\% and 68.31\% on the same datasets. The \model's strong performance on the MiniF2F theorem proving benchmark, which poses significant challenges due to its complexity and diversity, highlights the effectiveness of our proposed \method in enhancing the model's generalization ability.

The experimental results validate the effectiveness of \method in improving the performance of LLMs for autoformalization alignment evaluation. The integration of cross-entropy loss with contrastive learning in the model's training process has proven to be a powerful combination, resulting in a robust model capable of achieving high alignment-selection, precision, and recall scores across various datasets. The model's ability to generalize its performance to challenging benchmarks like MiniF2F further underscores the benefits of our approach.

\subsection{Comparison with Human Evaluation and LLM-As-Judge}

To comprehensively assess our \method model in autoformalization alignment evaluation, we conduct an extensive human evaluation along with an LLM-as-judge evaluation and compared their correctness rates. This analysis offers an in-depth understanding of our automated evaluation method's performance compared to human experts and state-of-the-art language models.

The experiment design, statistical results, as well as detailed discussions are specified in Appendix \ref{app:comprehensive_evaluation_and_manual_review}. As listed in \ref{tab: human_eval}, human experts achieved the highest correctness ratio in matching with the ground-truth alignment evaluations with an average of 79.58\%, followed by our~\method (65.00\%). The LLM-as-judge method achieves the lowest precision in autoformalization alignment evaluation. Each human expert takes approximately 3 hours to review 80 items, while the~\method model requires less than 2 minutes to conduct the automated evaluation. These findings emphasize the value of our \method framework in providing an efficient and reliable automated evaluation method for autoformalization alignment.
\section{Analysis and Discussion}
\label{sec:analysis}



To further validate the robustness and effectiveness of our \method framework, we conducted seven additional experiments, some of which are detailed in the Appendix due to limited space. We begin by validating the generalized effect of our \method across different baseline language models (\Cref{sec:exp_effect_of_different_base_llms}).
We investigate the necessity and effectiveness of our combined training loss (\Cref{sec:exp_effect_of_different_training_loss}) and the impact of our proposed alignment score \(\mathcal{V}_{\text{align}}\) (\Cref{sec:exp_effect_of_different_alignment_score}). 

Furthermore, we address concerns regarding potential data contamination in pre-trained language models through a comprehensive analysis of our experimental data (\Cref{app:data_contamination_analysis}). 
Next, we investigate the generalization ability of our method and the impact of different training datasets on its performance (\Cref{app:generalization_analysis}). 
We then explore the effect of incorporating contrastive learning loss on the performance of autoformalization of natural language statements to formal language statements (\Cref{app:autoformalization_performance_analysis}). 
To ensure the comprehensiveness and true representation of potential misalignments, we conduct an extensive manual review and evaluation of our \method framework (\Cref{app:comprehensive_evaluation_and_manual_review}).

These experiments collectively demonstrate the robustness, generalization ability, and effectiveness of our \method framework in various settings and highlight its potential for wider applicability in the field of autoformalization alignment evaluation.

\subsection{Effects of Different Baselines}
\label{sec:exp_effect_of_different_base_llms}
In this section, we validate the generalized effect of our \method across different baseline language models. 
These baselines are  Phi2-2.7B~\citep{javaheripi2023phi2} (\textbf{Phi}), LLaMA2-7B~\citep{llama2} (\textbf{LLaMA}), DeepSeekMath-Base 7B~\citep{Shao2024Deepseek-math} (\textbf{DeepSeek}) and Mistral-7B~\citep{Albert2023Mistral} (\textbf{Mistral}). 
\Cref{tab:baselines} presents the Alignment-Selection performance of the different baseline models across four datasets:




\begin{table}[h]
\centering
\caption{Alignment selection performance of different baselines across 4 datasets.}
\vspace{3pt}
\resizebox{0.5\textwidth}{!}{
\begin{tabular}{lcccc}
\toprule
\multirow{2}{*}{\textbf{Datasets}}    & \multicolumn{2}{c}{\textbf{\forml} } & \multicolumn{2}{c}{\textbf{MiniF2F} }   \\
\cmidrule(lr){2-3} \cmidrule(lr){4-5} & \textbf{Basic} & \textbf{Random} & \textbf{Valid} & \textbf{Test} \\
\midrule
\textbf{Phi}   & 80.77 &  71.07 & 31.56 &  32.51  \\
\textbf{DeepSeek}   & 90.29  &  77.08 &  54.66  &  55.19 \\
\textbf{LLaMA}    & 98.08  & 76.42 & 54.51  &  57.20  \\
\textbf{Mistral}   & 99.21 &  85.85  & 66.39  &  66.70 \\
\bottomrule
\end{tabular}
}
\label{tab:baselines}
\end{table}

The experimental results indicate that the Mistral model outperforms the other baseline models across all datasets, demonstrating the highest Alignment-Selection performance. The LLaMA and DeepSeek models perform strongly, particularly on the \forml datasets.
We note that the Phi model still performs adequately on the \forml datasets but struggles on the MiniF2F datasets, highlighting that our method is easily applicable to smaller models, as Phi2 has less than half the parameters compared to the other three models. These results validate the effectiveness of our \method in improving the automated alignment evaluation performance across various baseline language models, with Mistral showing the most significant improvements. This suggests that our \method can generalize well across different model architectures.

\subsection{Effects of Different Training Loss}
\label{sec:exp_effect_of_different_training_loss}


We investigate the necessity and effectiveness of our combined training loss, defined in Eq.~\eqref{eq:main_loss}, by conducting an ablation study with different loss configurations. The results, presented in~\Cref{tab:loss}, provide valuable insights into the impact of each loss component on the model's performance.


\begin{table}[h]
\centering
\caption{Comparison of overall alignment-selection performance across different configurations: with only cross-entropy loss (w/ CE), with only contrastive loss (w/ CL), and the complete model (Ours).}
\vspace{3pt}
\begin{tabular}{lcccc}
\toprule
\multirow{2}{*}{\textbf{Datasets}} & \multicolumn{2}{c}{\textbf{\forml}} & \multicolumn{2}{c}{\textbf{MiniF2F}} \\
\cmidrule(lr){2-3} \cmidrule(lr){4-5} & \textbf{Basic} & \textbf{Random} & \textbf{Valid} & \textbf{Test} \\
\midrule
\textbf{w/ CE} & 98.64 & 82.81 & 52.45 & 54.32 \\
\textbf{w/ CL} & 59.05 & 57.55 & 36.07 &  30.86 \\
\textbf{Ours}  & \textbf{99.21} &  \textbf{85.85}  & \textbf{66.39}  &  \textbf{66.70} \\
\bottomrule
\end{tabular}
\label{tab:loss}
\end{table}



\noindent \textbf{Autoformalization Inherently Learns Alignment:}The configuration using only the cross-entropy loss (\textbf{w/ CE}) achieves comparable performance, particularly on the \forml dataset. This result suggests that the autoformalization task, optimized by the cross-entropy loss, inherently learns alignment between informal and formal sequences. 


\noindent \textbf{Complementary Role of Contrastive Loss:}Although the configuration using only the contrastive loss (\textbf{w/ CL}) shows limited performance, it plays a crucial complementary role to the cross-entropy loss. The combined approach (\textbf{Ours}), which incorporates both cross-entropy and contrastive losses, achieves the best performance across all datasets. 
The combined loss function ensures that the \model benefits from both the sequence alignment inherent in the autoformalization process and the representation alignment facilitated by the contrastive learning process. By leveraging the strengths of both loss components, our model achieves a more holistic understanding of the task, enabling it to generate high-quality formal sequences that accurately capture the meaning and structure of the informal inputs.

\subsection{Effects of Different Alignment Score \texorpdfstring{$\mathcal{V}_{\text{align}}$}{Lg}}
\label{sec:exp_effect_of_different_alignment_score}
We investigate the necessity of our proposed alignment score \(\mathcal{V}_{\text{align}}\) as described in Eq.~\eqref{eq: alignment_score}. Table~\ref{tab:alignment-score-ablation} provides a comprehensive evaluation of the effectiveness of our proposed alignment score \(\mathcal{V}_{\text{align}}\). By analyzing different configurations of the model, we derive the following key insights:


\begin{table}[h]
\centering
\caption{Comparison of overall alignment-selection performance across: with only the certainty score (w/ cer), with only the similarity score (w/ sim), and the complete model (Ours).}
\vspace{3pt}
\begin{tabular}{lcccc}
\toprule
\multirow{2}{*}{\textbf{Datasets}} & \multicolumn{2}{c}{\textbf{\forml}} & \multicolumn{2}{c}{\textbf{MiniF2F}} \\
\cmidrule(lr){2-3} \cmidrule(lr){4-5} & \textbf{Basic} & \textbf{Random} & \textbf{Valid} & \textbf{Test} \\
\midrule
\textbf{w/ cer} & 98.98 &  85.64 &  53.69 & 55.55 \\
\textbf{w/ sim} & 45.25 & 20.75 & 20.49 &  21.81 \\
\textbf{Ours}  & \textbf{99.21} &  \textbf{85.85}  & \textbf{66.39}  &  \textbf{66.70} \\
\bottomrule
\end{tabular}
\label{tab:alignment-score-ablation}
\end{table}

\noindent \textbf{Language Generation Capabilities Build a Strong Basis:} The configuration using only the certainty score (\textbf{w/ cer}) achieves high performance, particularly on the \forml dataset. This result indicates that the model's language generation capabilities are robust and significantly contribute to the alignment evaluation. The certainty score measures the model's confidence in predicting the formal output, underscoring the importance of accurate language generation in our method.

\noindent \textbf{Superiority of Combined Score: } While using only the similarity score (\textbf{w/ sim}) shows limited performance, the combined approach (\textbf{Ours}), which integrates both certainty and similarity scores, achieves the best result across all datasets. This result demonstrates that combining both scores provides a more holistic and reliable evaluation metric. The combined score captures language-based and representation-level information, ensuring a robust evaluation during inference. 

In summary, the ablation study confirms that the combination of certainty and similarity scores provides a more robust and reliable metric for alignment evaluation. This integrated approach ensures that the evaluation metric reflects both the model's confidence in the generated outputs and the semantic alignment of the sequences, leading to superior performance in the AAE task.

\section{Conclusion}
\label{sec:conclusion}


In this study, we introduce \method, a framework designed to automate the alignment evaluation in the autoformalization process using LLMs. 
Our approach utilizes a dual loss function that combines cross-entropy and contrastive learning loss, significantly enhancing the model's ability to discern and align informal-formal language pairs. This methodology not only preserves the integrity of logical constructs but also improves the accuracy of alignment between informal and formal sequences.
Extensive experiments conducted across four datasets demonstrate that \method effectively reduces the reliance on manual verification processes, thereby streamlining the autoformalization workflow. 
The results confirm that our method provides reliable, effective, and robust evaluations, proving its practical utility in real-world scenarios.
We believe that \method opens new avenues for research and application in the autoformalization field, offering a scalable and efficient solution to one of the most pressing challenges in the domain.




\bibliography{iclr2025_conference}
\bibliographystyle{iclr2025_conference}

\appendix

\section{Lean 4 Compiler} 
\label{app:compiler}

The Lean 4 Compiler is a critical component of the Lean 4 programming language. This tool enables users to craft effective proof automation tactics within the Lean environment and transform them into optimized C code. The Lean 4 Compiler in our scope is referred to as the tool available at \url{https://github.com/leanprover-community/repl}. This particular resource provides a read-eval-print loop (REPL) designed for Lean 4, which supports user interaction through JSON formatted input and output streams (stdin and stdout, respectively). Our compilation projection is therefore founded on REPL. 

\section{More Related Works}

\paragraph{Formal Mathematics}

Formal mathematics has seen significant advancements with the development of interactive theorem provers (ITPs) such as Isabelle~\citep{MakariusIsabelle2008}, Lean~\citep{LeonardoLean42021}, HOL Light~\citep{Harrison1996Hollight}, and Coq~\citep{Barras1997Coq}. These systems serve as programming languages that allow users to input mathematical statements and proofs in a formal language for automatic verification. The field of autoformalization has emerged to bridge the gap between natural language mathematics and these formal languages~\citep{WangKU18,Szegedy20}, aiming to leverage the strengths of both: the extensive logical reasoning and human knowledge embedded in natural language~\citep{Lu2024YODA,Lu2024AUTOCV}, and the rigorous verification capabilities of formal systems~\citep{Kaliszy2014Corr}.

Recent years have witnessed substantial progress in dataset creation for formal mathematics. These efforts have primarily focused on two approaches: extracting information from established formal libraries and manually annotating or formalizing problems expressed in natural language. In the realm of data extraction, several datasets have been developed for popular proof assistants. Notable examples include LeanDojo~\citep{YangSGCSYGPA23} and MLFMF~\citep{BauerPT23}, which utilize the mathlib library in Lean. LeanDojo, in particular, has extracted an impressive collection of over 98,000 theorems and proofs, along with 130,000 premises from Mathlib.

Complementing these extraction efforts, researchers have also focused on manually formalizing problems from various mathematical domains. MiniF2F~\citep{ZhengHP22} stands out in this category, offering 488 manually formalized Olympiad-level problems across four proof systems, equally divided into validation and test sets. Other notable contributions include FIMO~\citep{LiuFIMO2023} and ProofNet~\citep{AzerbayevProofNet2023}, which formalize theorem statements from IMO and undergraduate-level problems in Lean.

Domain-specific formalizations have also gained attention. TRIGO~\citep{XiongTrigo2023} focuses on formalizing trigonometric reduction problems, while UniGeo~\citep{ChenLQLLCL22} and FormalGeo~\citep{Zhang2023FormalGeoTF} concentrate on annotating proof steps for geometry proving problems. These diverse datasets provide invaluable resources for researchers working on automated theorem proving, proof verification, and natural language processing in the context of formal mathematics. By offering a rich array of formalized mathematical content, they facilitate the development and evaluation of algorithms that can bridge the gap between natural and formal mathematical languages, potentially revolutionizing how we approach mathematical reasoning and verification in the digital age.

\section{Experimental Details}
\label{app:experimental-details}
\subsection{Finetuning Details}
Our experiments are conducted in a computing environment with 8 NVIDIA A100 GPUs, each with 40GB of memory. All models are fine-tuned in a full-parameter setting. 

We employ the AdamW optimizer for model training over 1 epoch, with a batch size of 512. The learning rate is set at \(2e \times 10^{-6}\), incorporating a 3\% learning rate warmup period. Below, we present a comprehensive overview of the training hyperparameters utilized. These parameters are consistently applied across training all LLMs.

\begin{table}[ht]
\centering
\caption{Finetuning Hyperparameters.~\label{tab:training-parameters-col-verifierer}}
\vspace{3pt}
\begin{tabular}{cc}
\toprule
\textbf{Hyperparameter} & \textbf{Value} \\
\midrule
Global Batch Size & 128 \\
LR & \(5 \times 10^{-6}\) \\
Epo. & 1 \\
Max Length & 2048 \\
Weight Decay & 0 \\
Warmup Ratio & 0.03 \\
\bottomrule
\end{tabular}
\end{table}

\subsection{Prompt Details}

\label{app:prompt-details}
We report greedy decoding results for GPT-4 and GPT-3.5 using a temperature setting of 0.0. 
Additionally, For the GPT-3.5 version, we query the API of gpt-3.5-turbo-0125. For GPT-4, we query the API of gpt-4-1106-preview.

\noindent\textbf{Prompt for Querying GPT for Automated Alignment Evaluation} Below, we provide the prompt used to query GPT for automated alignment evaluation. 
\begin{lstlisting}[style=informal]
Given an informal mathematical input and a formal theorem statement, your task is to evaluate the alignment between them. Assign a value between 1 and 5 to each formal output, where:

- 1 indicates that the formal output is not aligned with the informal input at all.
- 5 indicates that the formal output is perfectly aligned with the informal input. 

Consider the following criteria while assigning the values:

1. Semantic Consistency: How accurately does the formal output capture the meaning of the informal input?
2. Structural Correspondence: How well does the structure of the formal output reflect the structure implied in the informal input?
3. Completeness: Does the formal output include all relevant information from the informal input?
4. Precision: Is the formal output free from extraneous or incorrect information that is not present in the informal input?

Task:
1. Read the informal input.
2. Evaluate the formal theorem using the criteria above.
3. Assign a value between 1 and 5 to each formal output, reflecting its alignment with the informal input. Your output should follow this format: # Alignment Score: [your assigned value]

Informal Input:
{Informal_Input}

Pool of Formal Outputs:
{Formal_Outputs}

# Alignment Score:
\end{lstlisting}

We apply the prompt below for our \method model to obtain the alignment score without involving language generation settings.

\noindent\textbf{Prompt for Querying \method Model for Automated Alignment Evaluation:} Below, we provide the prompt used to query the \method model for automated alignment evaluation.

\begin{lstlisting}[style=informal]
Statement in natural language:

{Informal_Input}

Translate the statement in natural language to Lean:

{formal_output}   
\end{lstlisting}

\section{Data Contamination Analysis}
\label{app:data_contamination_analysis}

To address concerns regarding potential data contamination in pre-trained language models, we conducted a comprehensive analysis of our experimental data. This analysis is crucial, as language models are often trained on large amounts of unsupervised data, which may include samples similar to those used in our experiments.

\paragraph{Experiment Design}
We designed our experiments to mitigate the risk of data contamination by sourcing MiniF2F data from math olympiads such as AMC, AIME, and IMO. These datasets differ significantly from Mathlib, the largest Lean theorem library and the primary source of Lean data, reducing the likelihood of data contamination. Additionally, our automated alignment evaluation task involves augmenting aligned pairs with 20 negative examples using our proposed six misalignment strategies, ensuring that these data are not included in the pre-training corpus of LLMs.

\paragraph{Results}
We calculated the loss of different pre-trained models on the MiniF2F test/valid sets in the autoformalization task to further analyze the potential data contamination issue. 
This approach is inspired by the data contamination detection method in~\citep{Wei2023Skywork}, which suggests that if a language model has not been exposed to a dataset during pre-training, its loss on the dataset should be relatively high and approximately equivalent to its loss on a reference dataset composed of new, similar samples. The losses of the models in our experiments are shown below:

\begin{table}[htb]
\centering
\caption{Performance of Pre-trained Models on MiniF2F Datasets.~\label{tab:performance-pretrained-models}}
\vspace{3pt}
\begin{tabular}{lcc}
\toprule
\textbf{Pre-trained Model} & \textbf{MiniF2F-valid} & \textbf{MiniF2F-test} \\
\midrule
Phi2-2.7B & 2.4563 & 2.4377 \\
Mistral-7B & 1.4892 & 1.4660 \\
DeepSeekMath-Base 7B & 1.3148 & 1.2896 \\
LLaMA2-7B & 1.5343 & 1.5165 \\
\bottomrule
\end{tabular}
\end{table}

\paragraph{Analysis}
The loss values for each pre-trained model fall within the range of 1 to 3, consistent with and even higher than the findings in~\citep{Wei2023Skywork}, which reports that losses higher than around 1 on the GSM8K test set indicate low data leakage. These results suggest a low level of data contamination in our experimental data. The combination of carefully sourced datasets and the augmentation of aligned pairs with negative examples using our misalignment strategies further strengthens the robustness of our experiments against data contamination.

\section{Generalization Analysis}
\label{app:generalization_analysis}

To explore the generalization capabilities of our \method method, we conduct a series of experiments analyzing the impact of different datasets on the model's performance. These experiments aim to provide insights into the method's adaptability and effectiveness across various mathematical domains.

\paragraph{Experiment Design}

We design our experiments to assess the model's performance when trained on different datasets:
\begin{enumerate}
    \item  Our original model is fine-tuned on a combination of the FormL4 training set and the MMA training set from Mathlib.
    \item To evaluate the impact of individual datasets, we separately train models on FormL4 and MMA.
    \item We test all models on the MiniF2F test and valid sets, which are sourced from math olympiads such as AMC, AIME, and IMO, providing a fair comparison across challenging and diverse problem types.
\end{enumerate}

This approach allows us to gauge the generalization ability of our method and understand how different training datasets influence its performance.

\paragraph{Results}
The results of our experiments, focusing on the alignment-selection score for clear comparison, are presented in the following table:

\begin{table}[ht]
  \centering
\caption{Alignment-selection scores of different models on MiniF2F dataset.~\label{tab:generalization-analysis}}
\vspace{3pt}
  \begin{tabular}{lcc}
  \toprule\textbf{Model} & \textbf{MiniF2F Test} & \textbf{MiniF2F Valid} \\
  \midrule
  Ours (FormL4 + MMA) & 66.39 & 64.61 \\
  FormL4 only & 62.18 & 58.18  \\
  MMA only & 58.97 & 57.32 \\
  \bottomrule
  \end{tabular}
  \end{table}

Our results highlight several key findings:

\textbf{Dataset Content Impact}: TThe FormL4 dataset, which contains both statements and proofs, outperforms the MMA dataset, which only contains statements. This suggests that the inclusion of proofs provides richer information about the underlying mathematical concepts, leading to a more robust understanding of the alignment process.

\textbf{Synergy of Datasets:} Combining both FormL4 and MMA datasets for training results in improved performance compared to using either dataset alone. This demonstrates the potential benefits of leveraging diverse data sources to enhance the model's capabilities.

\textbf{Generalization Ability:} The strong performance on MiniF2F sets, which contain problems from challenging domains like math olympiads, indicates that our method can effectively handle diverse and complex mathematical problems. This suggests that \method has the potential for wider applicability across various mathematical domains.

These findings highlight the robustness of our \method method and its ability to generalize across different types of mathematical problems. The experiments demonstrate that by leveraging diverse datasets and considering both the quality and quantity of training data, we can enhance the method's performance and adaptability to new, unseen mathematical challenges.

\section{Autoformalization Performance Analysis}
\label{app:autoformalization_performance_analysis}
Given our model is primarily trained for the autoformalization task, we conduct additional experiments to explore its capabilities in converting natural language (NL) statements to formal language (FL) statements. These experiments aim to provide a comprehensive evaluation of our model's performance and demonstrate the effects of incorporating contrastive learning loss on autoformalization.

\paragraph{Experiment Design}
To assess the impact of contrastive learning loss on autoformalization performance, we compare two models:

1. A baseline model trained with cross-entropy loss only (\(\mathcal{L}_{CL}\))

2. Our proposed model, which incorporates both cross-entropy loss and contrastive learning loss (\(\mathcal{L}_{CL} + \mathcal{L}_{CE}\))

We evaluate both models on the FormL4 Basic and FormL4 Random test sets to obtain a comprehensive understanding of their autoformalization capabilities across different complexity levels.
The results of our comparison experiments are presented in the following table:

\begin{table}[ht]
\centering
\caption{Autoformalization performance of different models on FormL4 dataset.~\label{tab:autoformalization-analysis}}
\vspace{3pt}
\begin{tabular}{lcc}
\toprule
\textbf{Model} & \textbf{FormL4 Basic (\%)} & \textbf{FormL4 Random (\%)} \\
\midrule
Baseline (\(\mathcal{L}_{CL}\)) & 40.92 & 35.88 \\
Ours (\(\mathcal{L}_{CL} + \mathcal{L}_{CE}\)) & 43.14 & 36.02 \\
\bottomrule
\end{tabular}
\end{table}

\paragraph{Analysis}
The results demonstrate that incorporating contrastive learning loss improves autoformalization performance on both test sets. This improvement can be attributed to several factors:
\textbf{Enhanced Discrimination:} Contrastive learning acts as a form of data augmentation, introducing additional negative examples that enhance the model's ability to distinguish between correct and incorrect formalizations.

\textbf{Improved Representation Learning:} The contrastive approach helps the model learn more robust and discriminative representations of mathematical concepts, leading to more accurate autoformalization results.

\textbf{Generalization Across Complexity:} The performance improvement is observed in both the Basic and Random test sets, suggesting that the benefits of contrastive learning extend to various levels of problem complexity.

These findings highlight the potential of contrastive learning in improving autoformalization performance. By leveraging this approach, we not only enhance our model's capabilities but also pave the way for future research in this area. The success of incorporating contrastive learning loss suggests promising directions for developing more effective autoformalization techniques and advancing the field of automated mathematical reasoning.

Our experiments demonstrate that combining traditional cross-entropy loss with contrastive learning leads to a more robust and accurate autoformalization model. This approach could inspire further innovations in the field, potentially leading to even more sophisticated methods for bridging the gap between natural language mathematics and formal mathematical representations.

\section{Comparison with Human Evaluation and LLM-as-judge}
\label{app:comprehensive_evaluation_and_manual_review}

\paragraph{Experiment Design}

We design our experiment as follows:

\begin{enumerate}
    \item \textbf{Sample Selection:} We sample 80 items from the MiniF2F test set in our dataset. Originally, each item consists of:
    \begin{itemize}
        \item An informal natural language problem
        \item A formal statement
        \item A ground-truth label indicating alignment or misalignment between informal and formal statements
        \item The misalignment type (if the formal statement is misaligned with the informal one)
    \end{itemize}
    
    \item \textbf{Sample Distribution:} We ensure a balanced distribution between misalignment and alignment labels and include a diversity of misalignment types for a robust and representative evaluation.
    
    \item \textbf{Human Evaluation:} The same informal and formal statements in the 80 samples are provided to four human experts in Lean 4, who are tasked to independently evaluate autoformalization alignment (i.e., binary classification of alignment/misalignment).
    
    \item \textbf{Performance Metrics:} We calculate the correctness ratio of each human evaluator by comparing their assessments with the ground-truth labels.
\end{enumerate}

We similarly calculated the correctness ratio of our ~\method model by comparing its alignment selection results with the ground-truth labels (i.e., aligned/misaligned). The performance of GPT-4o, a state-of-the-art language model in LLM-as-judge research, was also obtained on the same task as our automated baseline. We used a scoring method with the instruction prompt provided in \ref{app:prompt-details} and searched for the best threshold to optimize the final correctness ratio.

\paragraph{Results}

The correct ratio (i.e., total percentage of the alignment evaluation results matching ground-truth labels) of GPT-4, ~\method model, and four human experts are listed below:


\begin{table}[ht]
\label{tab: human_eval}
\centering
\caption{Correctness ratio and agreement statistics of different evaluation methods on sampled MiniF2F test set.~\label{tab:manual-review}}
\vspace{3pt}
\begin{tabular}{lc}
\toprule
\textbf{Evaluation Method} & \textbf{Correct Ratio (\%)} \\
\midrule
GPT-4o & 47.50\% \\
\midrule
\method & 65.00\% \\
\midrule
Human Expert 1 & 83.75\% \\
Human Expert 2 & 77.50\% \\
Human Expert 3 & 77.50\% \\
Human Expert Average & 79.58\% \\
Fleiss' K & 0.49 \\
\bottomrule
\end{tabular}
\end{table}

As shown, human experts evaluation achieved the highest correctness ratio in matching with the ground-truth alignment evaluations with an average of 79.58\%, followed by our~\method (65.00\%). The LLM-as-judge method achieves the lowest precision in autoformalization alignment evaluation. Each human expert takes approximately 3 hours to review 80 items, while the~\method model requires less than 2 minutes to conduct the automated evaluation.


Our findings reveal several important insights:

\textbf{Efficiency and Robustness of \method:} Our \method framework provides a valuable automated method for evaluating autoformalization alignment due to its efficiency, robustness, and comparable accuracy. \method achieved a correctness ratio of 65.00\%, which is significantly higher than that of GPT-4o (47.50\%). With scaling, we believe that our automated method \method is promising to be even on par with the performance of human experts while requiring significantly less time for evaluation.

\textbf{Subjectivity of Manual Review:}  Manual review is subjectively dependent on the experts' domain knowledge and does not always achieve high accuracy or consistency. Notably, the human experts only reached a moderate interrater agreement ratio of 0.49. This highlights potential variability and inconsistency among the experts' evaluations.

\textbf{Complementary Role of Automated Evaluation:} The results underscore the need for automated evaluation methods to complement human reviews and ensure more consistent and objective alignment assessments. By leveraging the strengths of both manual and automated approaches, we can achieve a more comprehensive and reliable evaluation of autoformalization alignment.


The experiment also highlights the potential for further research in improving automated evaluation methods, as well as investigating the authentic representations of potential misalignments through detailed misalignment type analysis.

\section{Case Study}
\label{app:case_study}
We present a case study of a randomly selected informal-formal statement from our test dataset. We compare how our method and three other metrics (BLEU, BERTscore, Lean 4 Compiler) evaluate the alignment of various types of incorrect formal statements.
\begin{table}[ht]
\centering
\caption{Case Study: Comparison of Alignment Scores among misalignment types. Each evaluated formal statement is misaligned differently, as summarized in the table. All misaligned statements pass the Lean 4 Compiler without errors.}
\vspace{3pt}
\begin{adjustbox}{width=0.6\textwidth}
\begin{tabular}{lccc}
\toprule
\footnotesize
\textbf{Misalign type} & \textbf{\method} & \textbf{BLEU} & \textbf{BERTscore} \\
\midrule
Missing conditions & 0.56 & 0.82 & 0.98 \\
 Wrong Constant & 0.57 & 0.95 & 1.00 \\
Variable Type & 0.56 & 0.95 & 1.00 \\
Equality & 0.55 & 0.95 & 1.00 \\
Unpaired Statement & 0.57 & 0.12 & 0.90 \\
\bottomrule
\end{tabular}
\end{adjustbox}
\label{tab:accuracy_scores}
\end{table}








\begin{table}[ht]
\centering
\caption{Case Study: Visualized Examples of Misaligned Formal Statements.}
\vspace{3pt}
\begin{tabular}{p{0.95\textwidth}}
\toprule
\multicolumn{1}{c}{\textbf{Natural Language (Informal) Statement}} \\
\vspace{0.1em}
Prove that if $x \neq 0$ , $2x = 5y$ , and $7y = 10z$ , then $z/x = 7/25$ . \\
\midrule
\multicolumn{1}{c}{\textbf{Misaligned Formal Statements}} \\
\midrule
\begin{tabular}{p{0.3\textwidth}p{0.3\textwidth}p{0.3\textwidth}}
\begin{lstlisting}[style=lean]
theorem mathd_algebra_33 
  (x y z : ℝ) 
  (h₀ : 2 * x = 5 * y) 
  (h₁ : 7 * y = 10 * z) :
  Z / x = 7 / 25 := 
\end{lstlisting} &
\begin{lstlisting}[style=lean]
theorem mathd_algebra_33 
  (x y z : ℝ) 
  (h₀ : x ≠ 0) 
  (h₁ : 2 * x = 8 * y) 
  (h₂ : 7 * y = 10 * z) : 
  Z / x = 7 / 25 := 
\end{lstlisting} &
\begin{lstlisting}[style=lean]
theorem mathd_algebra_33 
  (x y z : ℚ) 
  (h₀ : x ≠ 0) 
  (h₁ : 2 * x = 5 * y) 
  (h₂ : 7 * y = 10 * z) : 
  Z / x = 7 / 25 := 
\end{lstlisting} \\
\end{tabular} \\

\midrule
\begin{tabular}{p{0.3\textwidth}p{0.3\textwidth}p{0.3\textwidth}}
\begin{lstlisting}[style=lean]
theorem mathd_algebra_33 
  (x y z : ℝ) 
  (h₀ : x = 0) 
  (h₁ : 2 * x = 5 * y) 
  (h₂ : 7 * y = 10 * z) : 
  Z / x = 7 / 25 := 
\end{lstlisting} & 
\begin{lstlisting}[style=lean]
theorem amc 12 b_2002_p 2 
  (x : ℤ) 
  (h₀ : x = 4) : 
  (3 * x - 2) * (4 * x + 1) - (3 * x - 2) * (4 * x) + 1 = 11 := 
\end{lstlisting}\\
\end{tabular}\\
\bottomrule
\end{tabular}
\label{tab:misaligned_statements}
\end{table}

Five types of misaligned formal statements are listed in Table \ref{tab:misaligned_statements}, together with the original natural language statements. As shown in~\Cref{tab:accuracy_scores}, for misalignments involving missing conditions, wrong constants, variable type mismatches, and equality violations, the \method scores are consistently below a threshold of 0.7, indicating low semantic precision of the formal statement and a likely misalignment. In contrast, both BLEU and BERTscore reported similarly high scores regarding various types of misalignment, demonstrating an inferior performance in evaluating the elusive misalignment in autoformalization.



\end{document}